# Narrative Context Protocol: An Open-Source Storytelling Framework for Generative AI

**Hank Gerba**
University of Southern California
Los Angeles, CA
Hg_119@usc.edu

### Abstract

Here we introduce Narrative Context Protocol (NCP), an open-source narrative standard designed to enable narrative interoperability, AI-driven authoring tools, real-time emergent narratives, and more. By encoding a story's structure in a "Storyform," which is a structured register of its narrative features, NCP enables narrative portability across systems as well as intent-based constraints for generative storytelling systems. We demonstrate the capabilities of NCP through a year-long experiment, during which an author used NCP and a custom authoring platform to create a playable, text-based experience based on her pre-existing novella. This experience is driven by generative AI, with unconstrained natural language input. NCP functions as a set of "guardrails" that allows the generative system to accommodate player agency while also ensuring that narrative context and coherence are maintained. The most up-to-date version of NCP can be found by accessing its GitHub repository: https://github.com/narrative-first/narrative-context-protocol/tree/main.

## Introduction

In their survey of authoring tools for computational narrative, Kybartas and Bidarra (2017) note that "we believe that creating a standard model of computational narrative could allow different systems to interact with the same narrative, without being restricted to incompatible models and definitions. Furthermore, such a model would also facilitate research into the generation of specific story components, e.g., allowing for multiple generators and even authors to collaborate on a given narrative." This paper proposes such a standard: Narrative Context Protocol (NCP).

Recent work has demonstrated that generative AI will enable a new paradigm of storytelling technologies and processes: from assisting a writer of linear media (novels, films, television, etc.) by allowing them to test out scenes and characters before committing them to a script, all the way through to real-time storytelling systems in videogames and other interactive media which respond to a player's agency, and countless use cases in between (Peng et al. 2024; Ranade et al. 2022). NCP is designed to serve any use case in which coherent narrative structure is a consideration, and in which authorial intent and direction is privileged. In the last five years, a robust body of research has demonstrated a variety of potential uses for computational narrative systems powered by generative AI, and some limited commercial deployments already exist (Yang et al. 2024; and Hu et al. 2024). With such promise, however, comes a series of challenges: technical, narrative, and ethical.

NCP was produced as an open and extensible standard. The ultimate directive of the project was to privilege author-centric design and functionality, enabling generative workflows which extend an author's narrative intent and creativity, rather than eclipse or replace it. NCP is an open-source method for encoding narrative structure in a way that is useful to authors and agnostic to the downstream narrative design applications in which it is deployed.

NCP's narrative model is based on the Dramatica theory of story, a framework created by Chris Huntley and Melanie Anne Phillips in 1994, and which has been used by screenwriters, authors, and other storytellers (Huntley and Phillips 2001). The model allows authors to capture their narrative intent in a "Storyform," a collection of interrelated key-value pairs which encode structural and temporal narrative logic. NCP is a structural answer to the problem of interactive narrative, offering a foundational thematic structure that ensures narrative coherence, while remaining flexible enough to be incorporated into any narrative design platform. Such narrative structures are particularly well-suited for applications using generative AI, as they act as a structured semantic reference which can serve as broad thematic "guard rails" for image, text, and audio generation. We envision that NCP-powered platforms will layer additional parameters onto both their authoring interfaces and story engines to shape a given experience according to each platform's specific needs.

A major concern with structural approaches to narrative is that they are unintuitive to authors, who feel constricted by the specific way in which a given structure encodes narrative information. Garbe (2020) has referred to this as both the complexity ceiling and authoring wall. We have been encouraged to find that workflows involving Large Language Models (LLMs) can ameliorate this problem by acting as an interpreter between the author and NCP's narrative structure. No matter an author's process, we have demonstrated that a properly configured LLM can be used to effectively translate natural language input into a properly configured NCP file. This greatly reduces the friction between authors and the narrative model, as they need only engage with the specifics of NCP's structure as needed or desired.

Because NCP operates at the level of narrative structure, it is extensible through supplementary systems such as plot managers, action planning mechanisms, etc. It is designed to provide meaningful narrative context, whether that means being deployed for real-time generative narrative, traditional writing aids, the management of serialized narratives, or any other use case which benefits from a rigorous narrative foundation (Kreminski and Martens 2022; Mirowski et al. 2023).

## Related Work

Throughout the 20th century, numerous structural approaches to narrative have been proposed and experimented with, rising with the advent of structuralist methodologies in the humanities. In his analysis of Russian fairy tales, Vladimir Propp (1968) proposed that stories comprise interlocking narrative "functions," the specific contents of which change in any given story. From a corpus of fairy tales, Propp formulated character functions as well as temporal functions. Propp's approach thus abstracted a set of semantic and syntactic rules which govern a well-formed story without saying anything of its specific content. In the early 1960's, linguist Richard E. Grimes worked with IBM to create a computational operationalization of Propp's fairy tales (Ryan 2017). The 1976 system TALE-SPIN (Meehan 1977) algorithmically generated stories based on rules derived from Aesop's Fables. Grasbon and Braun (2001) proposed a morphological approach to interactive storytelling based on Propp's model, writing that "We do not attempt to provide a model of generating stories in detail...Our primary concerns are high-level guidance of plot, as well as finding the best compromises between author and machine generation." Recent attempts to predict Proppian Functions from natural language have been successful (Valls-Vargas, Zhu, and Ontañon 2021). Riedl and Young (2010) have summarized these story generation techniques based on structural rules as s*imulation-based* narrative engines, which emergently produce a story by referencing an established narrative context governing features such as character motivation, plot progression, action sequences, and more. Several papers have summarized the history and current state of these models in much greater depth (Alhussain and Azmi 2022; Gallotta et al. 2024; Kybartas and Bidarra 2017; Ranade et al. 2022; Yang, Kleinmann, and Harteveld 2023).

While many such systems exist, each attempting to address various elements of what Mateas and Sengers (1999) have called narrative intelligence, or a generalized cognitive modeling particular to narrative, there has been no project whose primary aim has been to provide and maintain an open-source model of narrative structure available to downstream authoring platforms. Such a structure would allow for interoperability across authoring platforms, a common architecture for transmedial storytelling, the ability for multiple authors to work on the same story while retaining narrative context, and much more. We have chosen the Dramatica theory of story for this purpose because we believe it to be the most comprehensive model of narrative structure available, offering a structure which is extremely generalized, while at the same time offering fine-grained narrative control.

## NCP File Structure

NCP is an open, standardized JSON schema explicitly designed to transport and preserve authorial intent across diverse storytelling systems. At its core, NCP provides a structured yet adaptable schema, ensuring narratives retain their logical consistency and emotional depth, even when interpreted or extended by generative systems (Calderwood, Wardrip-Fruin, and Mateas 2022). By encoding narrative elements into explicit representations, NCP aims to maintain the original intent of the author through dynamic, distributed narrative environments.

The NCP schema separates narrative into two complementary layers: Narrative Structure and Storytelling.

- Narrative Structure represents the deeper, intended meaning of the narrative crafted by the author. It is represented by the Storyform, which is composed of Dynamics, Storypoints, and Storybeats.
- Storytelling is the adaptable, creative representation of this meaning to an audience.

Storytelling refers to the surface-level aspects of a narrative—the specific words, events, and details that an audience directly experiences. In traditional terms, this includes what most simply call the "story:" the concrete actions, dialogue, and characters presented explicitly within a narrative. For instance, William Shakespeare's *Romeo and Juliet* and the musical *West Side Story* differ significantly in Storytelling, yet share a nearly identical underlying Narrative Structure.

In contrast, Narrative Structure refers to the abstract, underlying logic that gives rise to meaningful storytelling. It operates beneath the surface, encoding subtextual meaning and thematic depth by structuring how inequities or conflicts are processed through multiple perspectives. NCP formally represents Narrative Structure in an object called a "Storyform." NCP is designed to make this formalization available so that it remains distinct from Storytelling, emphasizing that narrative meaning emerges not merely from the sequence of events in a story, but from the structured interplay of perspectives that interpret and respond to a set of underlying inequities.

Examples of complete NCP instances can be found at the projects GitHub repository.

## Narrative Structure

NCP builds upon the Dramatica theory of story, a narrative model developed by Chris Huntley and Melanie Anne Phillips in 1994. Designed to articulate narrative structures comprehensively, Dramatica provides a robust framework for narrative dynamics. In Dramatica's understanding, narrative is about the processing and resolution of inequity. Inequity does not necessarily imply overt conflict, rather an imbalance of thematic forces.

NCP contains the framework necessary to express a "Storyform," or the specific configuration of the Dramatica model which uniquely encodes the narrative intent of a particular story. Each element of the Storyform is interrelated, such that only certain configurations of the model are considered valid Storyforms. There is nothing that prevents an author from encoding a "broken" or "incomplete" Storyform, as in certain instances this may be desirable. Currently, the Subtxt/Dramatica narrative platform provides tools to generate validly configured Storyforms.

What follows is a high-level overview of a Storyform's structure, addressing both the spatial and temporal aspects which are required for a complete Storyform. Comprehensive details about Storyform structure are available at the Narrative Context Protocol (NCP) GitHub repository.

A Storyform consists of three main components: Dynamics, Storypoints, and Storybeats.

- Dynamics capture the broader strokes of what an author wishes to convey through their narrative, shaping the model toward the meaning intended for the audience.
- Storypoints represent specific sources of conflict, organized spatially as a series of nested quads.
- Storybeats are the temporal sequencing of events, emerging when Dynamics act upon Storypoints.

### Throughlines

At the core of every story lies an inequity, a fundamental imbalance between how things currently are and how they should or could be. This inequity represents a core tension compelling exploration and resolution, yet it defies complete understanding through any single perspective. Individually, each character struggles to fully grasp the imbalance because their perspective is inherently restricted to perceiving conflict from only one angle at a time, whether external or internal, static or dynamic.

As a complete structure, the narrative overcomes this limitation by simultaneously adopting multiple viewpoints, enabling those who experience it to appreciate the depth and complexity of the inequity. This multi-perspective structure makes it possible to explore the inequity's resolution, or understand the implications of leaving it unresolved.

For a Storyform to comprehensively explore its central inequity, it must integrate these four distinct Throughlines:

- Objective Story (OS): This is the Throughline through which we approach the various inequities of the story from a "They" perspective. The focus is on the external forces which shape the narrative's central inequity.
- Main Character (MC): This is the Throughline through which the audience or player approaches the central inequity of the story from an "I" perspective. It is often, but not always, coincident with the Protagonist—a particular configuration of dramatic functions, which, when consolidated in a single character, focus on pursuing the Story Goal.
- Influence Character (IC): This is the Throughline through which we approach the central inequity of the story from a "You" perspective. This perspective poses an alternative to the Main Character's sense of how to solve the narrative's inequity, which ends up catalyzing and provoking the Main Character to grow.
- Relationship Story (RS): This is the Throughline through which we approach the central inequity of the story from a "We" perspective. It concerns the growth of a relationship, often between the Main Character and the Influence Character, but it can also focus on another significant relationship within the story.

By weaving these four Throughlines together, a narrative fully explores an inequity in a way that we, as individuals, cannot. Each Throughline addresses the narrative's inequity in a unique way, leading to an emotionally fulfilling and narratively coherent conclusion.

### Players & Handoffs

Dramatica makes a distinction between "Characters" and "Players." Players correspond to what we typically think of as characters: specific people, creatures, etc. Players are characters from the perspective of the Objective Story—they represent the ensemble of characters within the narrative as a whole understood from the perspective of their specific collection of dramatic functions. This is where we find archetypal arrangements such as the Protagonist, Antagonist, and several others. Each archetypal Player is precisely defined by a specific configuration of dramatic elements. In the case of the Protagonist, one such dramatic function is the Motivation of Pursuit: they are the character who primarily represents the pursuit of the Story Goal. Of course, complex arrangements of dramatic functions are permissible, leading to Players with narrative complexity.

Additionally, each of the Throughlines may be "handed off" from one character to another. For instance, an author may shift the "I" perspective throughout the course of a story, handing off the Main Character Throughline from one player to another (*Game of Thrones).* Handoffs can occur in any of the Throughlines, such as when the challenge which the Influence Character represents to the Main Character is transferred from one Player to another. In such instances, the Player may change, but the thematic consistency and narrative progression of the story will remain coherent.

### Essential Storypoints: The Four Domains of Conflict

A Storyform is first shaped by aligning each Throughline with one of four essential Domains of conflict, establishing the foundational set of Storypoints.

- Universe: Static external conflict involving societal rules, status, finances, relationships, environment, or fixed limitations.
- Physics: Dynamic external conflict shown through actions, events, or visible changes.
- Psychology: Dynamic internal conflict involving psychological dysfunction, identity struggles, manipulative schemes, or distorted ways of thinking.
- Mind: Static internal conflict based on fixed beliefs.

Collectively, these four Domains attempt to encapsulate all possible manifestations of conflict, reflecting how our minds inherently process tension and discord, both externally and internally. Within each domain are further thematic elements: Types, Variations, and Elements (these are not discussed here for brevity). By applying these four Domains across the four Throughlines, a Storyform begins to concretely shape the narrative, ensuring a thorough and insightful exploration of the central inequity.

The four Througlines and Domains are spatialized according to a quad. There is a logic, expressed via the relative positions of terms within the quad, governing how each Throughline is associated with a Domain: the Main Character must always be positioned opposite the Influence Character, as they represent opposing personal viewpoints on resolving the central inequity. Similarly, the Relationship Story must be positioned opposite the Objective Story, as they represent opposing collective viewpoints—one subjective and relational, the other objective and external—thus ensuring a balanced exploration of the narrative's core conflict.

For instance, in a conventional Action Drama, the Objective Story is typically located in the Domain of Physics (*Star Wars, The Matrix, and Top Gun*), and the Main Character in Universe (Luke the farm boy, Neo "The One," Mav the son of Pete "Maverick" Mitchell). This reflects the fact that both the Main Character and the "big picture" of the story as a whole are concerned with navigating physical challenges. In a Courtroom Drama, by contrast, the Objective Story will typically be in Mind, with the Main Character operating in either Psychology (*A Few Good Men*) or Physics (*12 Angry Men*). This may reflect, for instance, that the overall inequity centers around pre-existing ideas about the innocence or guilt of a culprit, and that the Main Character attempts to go out and learn the real facts of the case (or manipulate those who stand in the way).

### Dynamics

Dynamics provide the broad strokes of an author's intent, shaping how a narrative's central conflict unfolds. Specifically, the Dynamic of Resolve conveys the author's underlying message by indicating whether the Main Character's Resolve is Relinquished or Maintained by the end of the narrative, particularly when faced with the contrasting perspective offered by the Influence Character.

When combined with the Dynamics of Outcome and Judgment, the author's core argument becomes clear. For example, a narrative may illustrate that relinquishing one's typical methods leads to a Successful outcome and resolves personal issues positively (Outcome: Success / Judgment: Good). Films such as *Star Wars* and *Top Gun* exemplify this thematic message.

Another dynamic is Problem-Solving Style. While films like *Star Wars* and *Top Gun* err more toward a Linear problem-solving style, others employ a holistic problem-solving style which is more about balancing forces than it is about solving a problem. *The Matrix* is one such example.

Finally, every narrative has a Limit Dynamic, which represents the way in which the ultimate point of tension is expressed. Narratives employ either a Timelock, in which a specific deadline necessitates the Main Character either Maintain or Relinquish their problem-solving method, or an Optionlock, in which the climax results from running out of options to avoid it.

There are several other important nuances to these Dynamics, such as whether actions drive decisions or decisions drive actions, which affect the algorithm through which Storypoints are temporally organized for thematic coherence. These dynamics can be further explored on the NCP GitHub repository.

### Storybeats

This complex interplay between various Dynamics and the sources of conflict encoded in the Storypoints results in an extremely varied temporal sequence of events across Sto-ryforms. The process by which NCP determines this temporal sequence is called Justification, aligning events according to the specific relationships between Storypoints and Dynamics within a given Storyform.

For example, if a Main Character's Dynamic of Resolve is Maintained, the progression of events may look like this:

- At the beginning, a force is introduced that challenges this character's worldview.
- Every event and decision within the story reinforces their commitment to this perspective.
- As the story escalates, pressure builds, leading to a final crisis where they must decide whether to stay the course or abandon their stance.
- The audience sees a pattern of persistence in the face of increasing opposition, culminating in a moment where either their resolve holds or their world collapses around them.
- In the end, this character chooses to stay the course, maintaining their resolve and fully embracing their perspective despite all opposition.

The Justification process yields a particular set of Storybeats, expressing the progression of narrative logic unique to a given Storyform. If the same characters were placed into a story where the Resolve is Relinquished, rather than Maintained, the NCP Justification process is capable of generating beat-by-beat temporal resolution reflecting this change. Similarly, changes to any other Dynamic or set of Storypoints will necessitate a renewed Justification to ensure the coherence of the narrative's logic.

While other narrative models generally do not provide temporal resolution beyond act structure, the NCP Justification process is capable of generating beat-by-beat temporal resolution.

## Storytelling Structure

NCP is designed to be human-readable. To make this simple, authors can associate storytelling details with any structural element. For instance, an author may add storytelling details to the Storybeats that roughly correspond to traditional act structure, or dive deeper and describe how each Dynamic, Storypoint, and Storybeat relates to their story. It is entirely up to the author how much storytelling detail they would like to include in each instance of NCP.

### Overviews

Overviews deliver high-level storytelling components, such as Throughline descriptions, plot summaries, and character arcs. These elements offer authors a clear understanding of the narrative's direction and key thematic drivers, supporting cohesive and engaging storytelling.

### Moments

Moments organize storytelling into narrative units like acts, scenes, chapters, or sequences. Each Moment includes a concise synopsis and structured references linking to associated Storybeats, providing clear narrative structure and aiding comprehension when passing NCP instances from one author to another.

## Interactive Narrative

Existing Dramatica theory has largely focused on non-interactive narratives. In this section we briefly highlight some of the capabilities NCP provides for interactive narrative.

### Player Modeling

In interactive narratives, the player is often cast as a character within the narrative. Pre-existing systems have attempted to deduce the narrative positioning of the player through their physical actions, dialogue choices, and other metrics (Ramirez and Bulitko 2014). Similarly, various attempts have been made to organize a "Theory of Mind" system undergirding Player and non-Player interactions (Chang and Soo 2008). With NCP, this process is simplified. For instance, all Players within the Overall Story are ascribed Elements within each of four Sets: Motivations, Methods of Evaluation, Purposes, and Methodologies.

There are 16 Elements in each Set, which are spatially arranged in a grid, for a total of 64 possible Elements. The spatial arrangement of these Elements is significant. For instance, Elements opposite one another in a given quad are considered Dynamic: they lead to the most dynamic narrative outcomes. A Motivation of Pursue, for instance, is opposite the Motivation of Avoid.

An archetypal Protagonist, embodies the Motivation of Pursue (they pursue the Story Goal), employs a Methodology of Pro-action, evaluates its progress by the Effect it has, and strives toward Actuality as its Purpose (bringing about the Story Goal).

Archetypal configurations are common, though do not come across as especially dimensional. For this reason, there is no limit on the Elements contained within a particular Player. This leads to complex and interesting characters.

In the context of an interactive narrative, a player's actions, dialogue, and other metrics can be tracked and associated with each of the four Sets, leading to a nuanced understanding of their problem-solving approach within the context of the narrative. Because the relationships between each Element of a Set are codified, this can also be used to generate narratively interesting moments by activating those Players whose construction is inherently in a dynamic relationship with that of the player.

### Recombinant Storyforms

NCP can be used to alter the Storyform in response to player agency, ensuring that a narratively coherent outcome always occurs. If, for instance, the player has largely pursued a Steadfast approach to their problem-solving, an NCP-driven experience may perform the Justification process with the Dynamic of Resolve set to Maintained, generating a Storyform which takes this narrative logic into account. However, if at the climax of the narrative the player decides to relinquish their approach and try something new, the Storyform governing the experience must be modified to yield a narratively satisfying conclusion. In an NCP-driven platform, the Justification process can be used to quickly recalculate the Storyform to reflect this new Dynamic, affecting the Storyform for the remainder of the story.

### Convolutional Storyform / Storyform Inheritance

Because the Justification process is a function, it is possible to develop interrelated, or convolutional, Storyforms. This allows many Storyforms to nest, overlap, spawn, and respawn dynamically depending on the evolution of the story. If, for instance, a multiplayer experience allows for player-generated stories, but maintains a broader narrative across all sessions, NCP could be used to manage the evolving relationship between these narrative strata.

### Semantic World Engine Managment

In a simulated environment, NCP can provide narratively significant semantic relationships between objects, characters, scenes, etc. This information can then be used by generative systems—visuals, dialogue generation, etc.--to enable narrative interactivity which has traditionally been prohibitively resource-intensive. For instance, a player could ask any non-player character (NPC) about any object in their inventory, at any particular point in the story, and receive narratively significant responses based on each elements relationship as encoded in the Storyform.

## Results

NCP is designed to be used in a wide variety of contexts, allowing for the transport of narrative information across domains. To demonstrate its capabilities as both an interpretive schema and template for translating narratives between mediums, a custom authoring tool was developed within the Dramatica/Subtxt Platform, which is powered by OpenAI's o3 and GPT-4o models. We are therefore experimenting with pre-trained models, which has proven successful in other studies (Chung et al. 2022; Guan et al. 2023). This tool has a number of capabilities:

- It allows an author to converse with an LLM, developing a story idea from something as simple as a logline through to a complete Storyform. This storyform can be exported as an NCP instance.
- Once a Storyform has been generated, each Moment can be described with Storytelling, according to traditional sequences such as acts, scenes, and beats.
- In "Play Mode," these Moments can be experienced from the perspective of any character within the narrative, or from the player's own perspective as a new character. This mode places no restriction on a player's natural language input, allowing them to dramatically alter the Storytelling. Because the Moment's narrative meaning has been encoded in NCP, the LLM is able to satisfyingly resolve the scene from the perspective of its intended narrative structure. NPCs are entirely driven by the LLM (Gao and Emami 2023).
- In "Director Mode," which is meant to simulate a future authoring paradigm for emergent narrative, Moments progress beat-by-beat, presenting the author with a range of narrative information about the current beat: actions taken by each character, their instantaneous motivation, dialogue, blocking, etc. At any moment the author can alter the behavior of any character, introduce new elements into the scene, etc. In this way the mode simulates a "stop-and-go" rehearsal.

NCP was used to assist an author in transforming her traditionally written novella into an interactive, text-based experience powered by generative AI. The goal was to use NCP to understand the narrative intent behind the novella, and supply it to an LLM-driven experience as a set of contextual guardrails. This ensured that no matter the player's input, and thus no matter the Storytelling, the narrative meaning of each scene could be steered toward the author's intended narrative structure.

Our author began by writing a science fiction novella based on Charlotte Brontë's *Jane Eyre*, titled *Jane Air.* During this process, our writer wrote as she normally does, with no consideration for any further development of the narrative. In this version of the story, Jane hails from a small moon on the edge of the solar system, and is hired as a ward upon Captain Rochester's trading vessel, The Thorn. The narrative broadly follows that of its source, but deviates significantly enough that, after analysis, its Storyform is slightly different.

By uploading the completed novella to the Subtxt/Dramatica Platform, a back-and-forth conversation between the author and the LLM arrived at a Storyform which felt true to the author's narrative intent. Once the Storyform has been deduced, it can either be operated on internally within the platform, or exported as an NCP file to be operated upon in any other authoring platform. This "mixed-initiate" approach has been deployed by others (Kreminski et al. 2022, Stefnisson and Thue 2018).

Experiences in Play mode are divided into Moments, into which the author can sort the various Storybeats which accord with a given scene in the novella. This approach mirrors other "storylet" based techniques (Kumaran, Rowe, and Lester 2024; Kumaran et al. 2023; Mason 2021). For instance, the novella contained a prologue, in which Jane is still a child living in a remote mining installation called Gates. Here she is watched over by the ruthlessly cruel Mother Reed, navigating diligently to avoid her watchful gaze. The scene ends when a ship flies in from off-world, providing Jane with a glimmer of hope for her future. According to the Storyform, the scene moves thematically from Storybeats of Certainty to Deduction, Deduction to Induction, and finally Induction to Potentiality.

In the platform, these Storybeats are associated with a Moment corresponding to the scene, so that the system has a thematic progression it can adhere to as the player makes their own Storytelling choices. This ensures that the scene contains a similar meaning to the corresponding scene in the novella, despite the fact that the player may radically alter the events that actually take place. It also has the added benefit of allowing the system to automatically end the scene naturally, once it recognizes that the final Storybeat has been explored.

It may be desirable for an author to introduce finer-grained controls over the scene. Because Play Mode is meant to demonstrate the capabilities of the system to manage narrative structure, we elected to provide only a few additional control points, with the expectation that authoring tools which use NCP will have much more specific use cases which demand a variety of controls. The one control method we did introduce is called "Imperatives." These are Storytelling moments which an author requires be present in every playthrough. For instance, in the Gates scene, the author introduced an imperative that Mother Reed must, at some point in the scene, say to Jane that "You'll never leave here." The point of this line is to underscore Jane's feeling of being trapped. When a user plays through the scene, the system works to naturally integrate this line of dialogue at some point during the scene.

In "Director Mode," we strove to replicate the structure of a stop-and-go rehearsal. In this mode, instead of text being presented novelistically, the system provides a high-level overview of everything that is happening in the scene. The scene progresses beat-by-beat, asking after each beat if the author would like to make alterations to the narrative content of the scene. Information presented includes listing the characters present, any actions they are about to take, the dialogue they deliver, their instantaneous motivation, a description of the environment, and more. These parameters are alterable, and because they all derive from the system's understanding of the narrative, any feature requested will be generated primarily to serve the story.

At any point, the author can interject changes to the story. For instance, at the beginning of the Gates scene, Mother Reed brusquely intrudes on Jane as she peruses a data pad filled with images of other worlds in her room. The author may decide that Mother Reed should be a bit more manipulative, knocking politely and behaving as if she has Jane's best interests at heart, all the while concealing her true nature. After describing this change, the system rewrites the intrusion beat with this revised characterization. In a more sophisticated version of this platform, these changes would then be saved into NCP characterization of Mother Reed, such that subsequent run-throughs of Play Mode would reflect this new vision of the character. In this version, the author manually made these changes to the character description based on whether or not they liked how the revised scene played out.

## Limitations & Future Work

Because NCP is a structural approach to narrative standardization, it requires a certain level of confidence that Storyforms are an adequate representation of narrative structure. Of course, this is an ineliminable consequence of proposing any specific narrative model as a standard. It is our belief that the Dramatica model of story is the most useful and feature-rich model of narrative available, as each feature within the model is rigorously defined, as are the specific relationships between these features.

This leads us to a second limitation, which is the present study's lack of standardized success metrics, as others have demonstrated (Purdy et al. 2018, and Partlan et al. 2018). While both Play Mode and Director Mode provide summaries upon completion of a scene which demonstrate how the model attempted to fit the Storytelling with the Narrative Structure for that scene, we did not have the ability to compare results over time in a rigorous way. In future work we plan to standardize a mode of evaluation which tracks the presence (or absence) of features such as Throughlines in a given narrative, as well as to what extent the Storytelling of a scene expresses its underlying Storybeats.

Finally, while we have sketched a number of capabilities NCP provides for interactive narrative, they remain untested. In future work, we hope to more rigorously theorize possibilities, as well as demonstrate them in practice.

Additionally, we plan to update NCP to serve as a ledger for use in copyrighting and the protection of intellectual property. We envision that authors could use NCP as a way of tracking contributions to a particular project, and in particular the ways in which their inputs shape generative narrative systems. Such a ledger-based system would help protect authors and ensure that attribution is properly maintained.

## Conclusion

In introducing the capabilities and features of NCP, and in providing some examples of its potential use, we demonstrate the utility of a standardized model of computational narrative. We are particularly sensitive to concerns surrounding generative AI, particularly concerns about labor and the displacement of writers. We have designed NCP so that it operates at the level of narrative structure, which, on its own, does not make for good storytelling. Of his 1960 computational modeling of Propp's model of narrative structure, Richard E. Grimes said that "The thing I never put my finger on was that my computer's stories had Propp's elements and sequences, but they were all boring" (Ryan 2017).

We take this to mean that storytelling requires human creativity to be meaningful—structure is not enough. NCP is designed to be a standard that leaves storytelling to authors, while providing tools to interface a new generation of authoring tools, and indeed to explore the capabilities of these technologies. The question of ethical use comes down primarily to specific application: when used to facilitate communication between authors and artists, enable new kinds of narrative experiences, or otherwise enrich creative output rather than streamline budgets, generative technologies can and will be interesting tools.

## Acknowledgments & Funding Statement

The Narrative Context Protocol Project was funded by a consortium led by the Entertainment Technology Center at the University of Southern California. The consortium comprised Amazon Studios, The Walt Disney Company, Dolby Laboratories Inc., Epic Studios, and NBCUniversal. Theoretical development was facilitated by Jim Hull and Paul Bennun.

## References

Alhussain, A. I.; and Azmi, A. M. 2022. Automatic Story Generation: A Survey of Approaches. ACM Computing Surveys, 54(5): 1–38.

Calderwood, A.; Wardrip-Fruin, N.; and Mateas, M. 2022. Spinning Coherent Interactive Fiction through Foundational Model Prompts. In *Proceedings of the 13th International Conference on Computational Creativity, Bozen-Bolzano, Italy, June 27-July 1, 2022,* 44-53. Association for Computational Creativity (ACC).

Chang, H. M.; and Soo, V. W. 2008 Simulation Based Generation with a Theory of Mind. *Proceedings of the AAAI Conference on Artificial Intelligence and Interactive Digital Entertainment*, 4(1), 16-21.

Chung, J. J. Y.; Kim, W.; Yoo, K. M.; Lee, H.; Adar, E.; and Chang, M. 2022. TaleBrush; Sketching Stories with Generative Pretrained Language Models. In *Proceedings of the 2022 CHI Conference on Human Factors in Computing Systems,* 1-19.

Gallotta, R.; Todd, G.; Zammit, M.; Earle, S.; Liapis, A.; Togelius, J.; and Yannakakis, G. N. 2024. IEEE Transactions on Games, 1-18.

Gao, Q. C.; and Emami, A. 2023. The Turing Quest: Can Transformers Make Good NPCs? In Padmakumar, V.; Vallejo, G.; and Fu, Y. (eds), *Proceedings of the 61st Annual Meeting of the Association for Computational Linguistics (Volume 4: Student Research Workshop)*, 93-103. Association for Computational Linguistics.

Garbe, J. 2020. Increasing Authorial Leverage in Generative Narrative Systems. PhD dissertation, UC Santa Cruz.

Grasbon, D.; and Braun, N. 2001. A Morphological Approach to Interactive Storytelling. In Proc. CAST01, Living in Mixed Realities. Special issue of Netzspannung, the Magazine for Media Production and Inter-media Research, 337-340.

Guan, L.; Valmeekam, K.; Sreedharan, S.; and Kambhampati, S. 2023. Leveraging Pre-trained Large Language Models to Construct and Utilize World Models for Model-based task planning. *Advances in Neural Information Processing Systems,* 36:79081-79094.

Hu, S.; Huang, T.; Ilhan, F.; Tekin, S.; Liu, G.; Kompella, R.; and Liu, L. 2024. A Survey on Large Language Model-based Game Agents, 2024. arXiv:2024.02039.

Huntley, C.; and Phillips, M. A.; 2001. *Dramatica, a New Theory of Story.* California: Screenplay Systems Incorporated.

Kreminski, M.; and Martens, C. 2022. Unmet Creativity Support Needs in Computationally Supported Creative Writing. In Proceedings of the First Workshop on Intelligent and Interactive Writing Assistants (In2Writing 2022), 74–82. Dublin, Ireland: Association for Computational Linguistics.

Kreminski, M.; Dickinson, M.; Wardrip-Fruin, N.; and Mateas, M. 2022. Loose Ends: a Mixed-initiative Creative interface for playful storytelling. In *Proceedings of the AAAI Conference on Artificial Intelligence and Interactive Digital Entertainment,* volume 18, 120-128.

Kumaran, V.; Rowe, J.; and Lester, J. 2024. NarrativeGenie: Generating Narrative Beats and Dynamic Storytelling with Large Language Models. *Proceedings of the AAAI Conference on Artificial Intelligence and Interactive Entertainment*, 20(1), 76-86.

Kumaran, V.; Rowe, J.; Mott, B.; and Lester, J. 2023. SCENECRAFT: Automating interactive Narrative Scene Generation in Digital Games with Large Language Models. In *Proceedings of the AAAI Conference on Artificial Intelligence and Interactive Digital Entertainment,* volume 19, 86-96.

Kybartas, B.; and Bidarra, R. 2017. A Survey on Story Generation Techniques for Authoring Computational Narratives. IEEE Transactions on Computational Intelligence and AI in Games, 9(3): 239-253.

Mason, S. 2021. Responsiveness in Narrative Systems. PhD dissertation, University of California Santa Cruz, Santa Cruz, CA.

Mateas, M.; and Sengers, P. 1999. Narrative Intelligence. In Mateas, M.; and Sengers, P. (eds), *Narrative Intelligence: Papers from the 1999 Fall Symposium (Technical Report FS-00-01)*, pp. 1-10. Menlo Park: AAAI Press.

Meehan, J. R. 1977. TALE-SPIN: An Interactive Program that Writes Stories. In *Proceedings of the 5th Joint Conference on Artificial Intelligence*, 91-98.

Mirowski, P.; Matheson, K. W.; Pittman, J.; and Evans, R. 2023. Co-Writing Screenplays and Theatre Scripts with Language Models: Evaluation by Industry Professionals. In *Proceedings of the 2023 CHI Conference on Human Factors in Computing Systems,* 1-34.

Partlan, N.; Carstensdottir, E.; Snodgrass, S.; Kleinman, E.; Smith, G.; Harteveld, C.; and Seif El-Nasr, M. 2018. Exploratory Automated Analysis of Structural Features of Interactive Narrative. *Proceedings of the AAAI Conference on Artificial Intelligence and Interactive Digital Entertainment,* 14(1), 88-94.

Peng, X.; Quaye, J.; Rao, S.; Xu, W.; Botchway, P.; Brockett, C.; Jojic, N.; DesGarennes, G.; Lobb, K.; Xu, M.; Leandro, J.; Jin, C.; and Dolan, B. 2024. Player-driven Emergence in Llm-driven game narrative. arXiv:2404.17027.

Propp, V. 1968. *Morphology of the Folktale.* Austin: University of Texas Press.

Purdy, C.; Wang, X.; He, L.; and Riedl, M. 2018. Predicting generated story quality with quantitative measures. In Proceedings of the AAAI Conference on Artificial Intelligence and Interactive Digital Entertainment, volume 14, 95–101.

Ramirez, A.; and Bulitko, V. 2014. Automated Planning and Player Modeling for Interactive Storytelling. IEEE Transactions on Computational Intelligence and AI in Games, 7(4): 375-386.

Ranade, P.; Dey, S.; Joshi, A.; Finin, T.; 2022. Computational Understanding of Narratives: a Survey. IEEE Access(10):101575-101594.

Riedl, M. O.; and Young, R. M. 2010. Narrative Planning: Balancing Plot and Character. Journal of Artificial Intelligence Research, 39: 217–268.

Ryan, J. 2017. Grimes' Fairy Tales: a 1960s story generator. In: Nunes, N.; Oakley, I.; Nisi, V., (eds); Interactive Storytelling. ICIDS 2017. Lecture Notes on Computer Science(), vol 10690.

Stefnisson, I.; and Thue, D. 2018. Mimisbrunnur: AI-assisted Authoring for Interactive Storytelling. In *Proceedings of the AAAI Conference on Artificial Intelligence and Interactive Digital Entertainment,* volume 14, 236-242.

Valls-Vargas, J.; Zhu, J.; and Ontañón, S. 2021. Predicting Proppian Narrative Functions from Stories in Natural Language. *Proceedings of the AAI Conference on Artificial Intelligence and Interactive Digital Entertainment,* 12(1), 107-113.

Yang, D.; Kleinman, E.; and Harteveld, C.; 2024. GPT for Games: A Scoping Review (2020-2023). In proceedings of the IEEE Conference on Games (COG).